\documentclass{article}

\usepackage[final, nonatbib]{neurips_wrl2021}

\usepackage[T1]{fontenc}
\usepackage[utf8]{inputenc}
\usepackage{graphicx}
\usepackage{amsmath}
\usepackage{amssymb}
\usepackage{float}
\usepackage{multirow}
\usepackage{array}
\usepackage[strict]{changepage}
\floatstyle{plaintop}
\restylefloat{table}
\usepackage{url}
\usepackage{hyperref}
\usepackage{microtype}
\usepackage{cite}
\usepackage{bm}
\usepackage{enumitem}  
\newcounter{alphasect}

\def\alphainsection{0}

\let\oldsection=\section
\def\section{%
  \ifnum\alphainsection=1%
    \addtocounter{alphasect}{1}
  \fi%
\oldsection}%
\renewcommand\thesection{%
 \ifnum\alphainsection=1% 
   \Alph{alphasect}%
 \else
   \arabic{section}%
 \fi%
}%

\usepackage[]{algorithm2e}

\usepackage{color}

\title{Assistive Tele-op: Leveraging Transformers to Collect Robotic Task Demonstrations}

\author{Henry M. Clever$^{1,2}$, Ankur Handa$^{1}$, Hammad Mazhar$^{1}$, Kevin Parker$^{1}$, Omer Shapira$^{1}$, \\ \textbf{Qian Wan$^{1}$, Yashraj Narang$^{1}$, Iretiayo Akinola$^{1}$, Maya Cakmak$^{1}$, Dieter Fox$^{1}$}  \\
  $^{1}$NVIDIA, USA. 
  $^{2}$Georgia Institute of Technology, Atlanta, GA, USA. \\
}

\begin{document}

\maketitle
%\vspace{-6mm}

\begin{abstract}
%\vspace{-4mm}
 Sharing autonomy between robots and human operators could facilitate data collection of robotic task demonstrations to continuously improve learned models. Yet, the means to communicate intent and reason about the future are disparate between humans and robots. We present Assistive Tele-op, a virtual reality (VR) system for collecting robot task demonstrations that displays an autonomous trajectory forecast to communicate the robot’s intent. As the robot moves, the user can switch between autonomous and manual control when desired.  This allows users to collect task demonstrations with both a high success rate and with greater ease than manual teleoperation systems. Our system is powered by transformers, which can provide a window of potential states and actions far into the future -- with almost no added computation time. A key insight is that human intent can be injected at any location within the transformer sequence if the user decides that the model-predicted actions are inappropriate. At every time step, the user can (1) do nothing and allow autonomous operation to continue while observing the robot’s future plan sequence, or (2) take over and momentarily prescribe a different set of actions to nudge the model back on track. We host the videos and other supplementary material at \url{https://sites.google.com/view/assistive-teleop}.
  
\end{abstract}

%\vspace{-6mm}
\section{Introduction}
\begin{figure*}[t!]
\centering
\includegraphics[width=14cm]{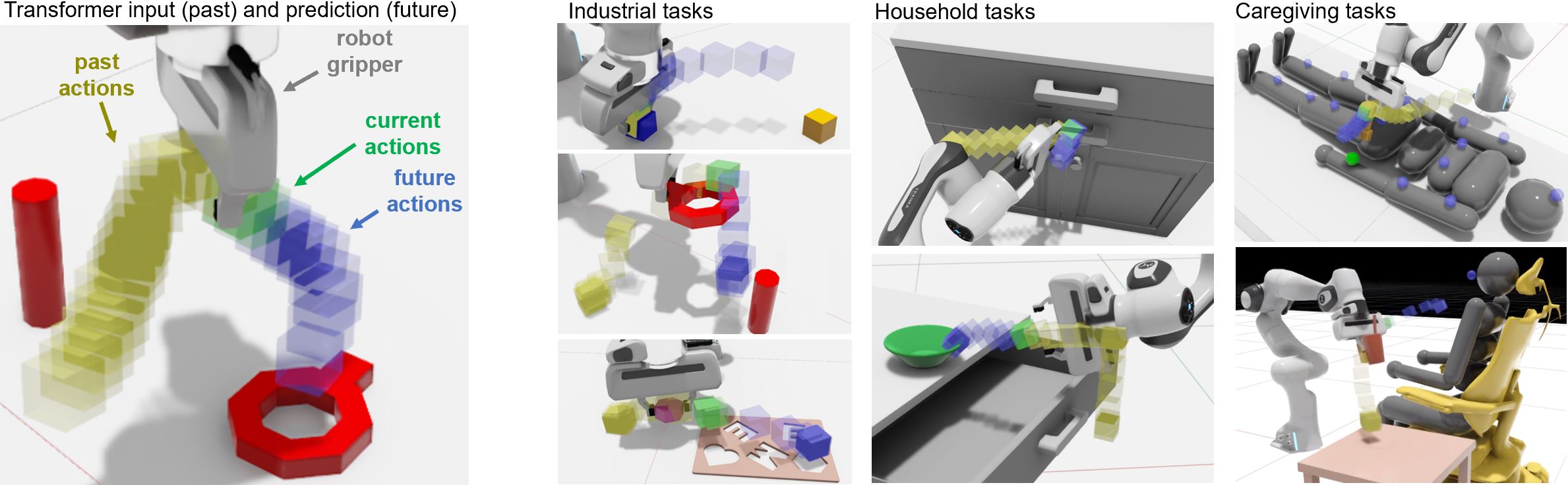}
\vspace{-5mm}
\caption{\footnotesize{Assistive Tele-op using transformers for data collection. \textit{Left:} Robot motion displaying input past actions (yellow) and output predicted actions (blue) to the user. \textit{Right:} We test our method on 7 task scenarios across industrial (block stacking, nut assembly, kit assembly), household (cabinet drawer and bowl manipulation), and caregiving (itch scratching and drinking) tasks.}}
\label{fig:overall_fig}
\vspace{-2mm}
\end{figure*}

%\vspace{-3mm}
\let\thefootnote\relax\footnotetext{E-mail: ahanda@nvidia.com, henryclever@gmail.com} Manually teleoperating robots to collect task demonstrations at scale is laborious and challenging. We present a shared-autonomy-based method using neural networks to forecast robot trajectories that substantially reduces manual teleoperation time while maintaining a high success rate. Specifically, we adapt a learned model to do trajectory auto-complete, \textit{i.e.,} given an initial sequence of states and actions, the network learns to complete the rest of the trajectory. The user can either accept the model's suggestions or provide manual corrections  while observing their effect on the model forecast. We leverage transformers~\cite{vaswani2017attention} for modeling the states and actions through time, which are well-suited to modeling long sequences of information with complex dependencies (see Fig.~\ref{fig:overall_fig}-\textit{left}). Their self-attention mechanism can holistically understand a robot trajectory, rather than emphasizing adjacent connections between states. When taking as input a sequence of past actions, the transformer can look far into the future and predict future actions. By integrating this into a robot manipulation environment with VR, a user can decide if executing the future actions is appropriate, or otherwise take momentary control of the system to provide a better demonstration.

This work explores transformer model prediction for a set of 7 manipulation tasks involving pick-and-place across industrial, household, and caregiving robot settings (Fig. 1-\textit{right}). We train on a large number of demonstrations ($>500$) from the open-source RoboTurk \cite{mandlekar2018roboturk} dataset and perform few-shot learning by fine-tuning on a small number of expert demonstrations ($<=60$) for each of the 7 tasks. We show that the model is able to succeed autonomously for 67.1\% of task scenarios during test time. When the model predicts a wrong sequence of actions, the user takes control and gives the robot a nudge to get it back on track. This significantly improves performance, resulting in a 96.1\% success rate. Importantly, our Assistive Tele-op system reduces manual control time to collect demonstrations by a factor of 5. It is worth stressing that while the use of interventions is similar to DAgger \cite{Ross:etal:2011}, we aid the user by displaying a live forecasting model of the robot trajectory so corrections can be made well before the robot makes a mistake.

% With our formulation into the transformer, we can map a variety of tasks into a single state space of fixed length. 
\textcolor{black}{Our transformer embedding has a single state space of fixed length that we can map a variety of tasks into. Further, by defining the robot state and actions entirely in end-effector space like~\cite{martin2019variable} and learning a world model~\cite{ha2018recurrent} in this space, the transformer can be interchanged} between robots and simulation environments. Our transformer is pre-trained on demonstrations collected in a different physics simulator and robot than what we use in this work. While only the predicted robot actions are necessary for control, we use an additional loss on object pose states in the scene, which boosts performance. Transformers are better able to parse sequence information using a positional embedding vector, which we compute using the cumulative distanced traversed by the end effector in Euclidean space (similar in spirit to~\cite{csordas2021devil}). Finally, by training with a BERT-style zero padding~\cite{devlin2018bert} on the input to the transformer corresponding to future states, we can make an arbitrary number of model predictions far into the future -- which allows the user to understand what the robot is about to do. % If pretrained on a large dataset, transformers have been known to generlize well to new tasks given few labeled examples~\cite{logeswaran2020few}. We explore this by pretraining our robotic transformer on a large ($>500$) set of pick-and-place task demonstrations in the open-source RoboTurk dataset~\cite{mandlekar2018roboturk}. 

In summary, the work makes the following contributions: (1) Evidence that pre-trained transformers can be fine-tuned for few-shot generalization to new robot manipulation tasks, and (2) Assistive Tele-op, a VR system with live model forecasting, to assist users with collecting robotic task demonstrations at a high success rate \textcolor{black}{and with substantially reduced manual control time. As the user collects more demonstrations, they can be fed back to the model for continual learning.} % with substantially reduced manual effort. 

    %\item Pre-training a transformer model on a large dataset of demonstrations collected from the RoboTurk dataset \cite{mandlekar2018roboturk}, across four pick-and-place tasks. The tasks can be mapped to a state space of same size enabling training multiple different tasks together.
    
    %\item Fine-tuning the model on 7 new tasks in a different simulator using a handful of demonstrations and showing that the performance is better than training from scratch. 
    
    %\item Since the model is trained to make trajectory predictions far into the future, they can be displayed live in VR to assist teleoperation users with collecting robot demonstrations. 
    
    %\item We also use live forecasting to make slight corrections to the model predictions that deviate from the expected path further improving the task performance significantly.
%\end{itemize}

%\begin{figure}[t!]
%\centering
%\includegraphics[width=5cm]{pics/transformer_state_formulation.PNG}
%\vspace{-3mm}
%\caption{The input and output to the transformer consist of robot states and actions in end effector space, as well as the pose of objects in the scene. The action is a target end effector pose that may or may not be reached at any point in time.}
%\vspace{-3mm}
%\label{fig:transformer_state}
%\end{figure}

%\vspace{-3mm}
\section{Related Work}\label{sec:related_work}
Task demonstrations for robot manipulation can be collected using automatic methods such as trajectory optimization~\cite{bryson2018applied, levine2014learning} and reinforcement learning~\cite{schulman2017proximal, lillicrap2015continuous, Ross:etal:2011}, as well as manual methods of kinesthetic teaching and teleoperation~\cite{argall2009survey}. %The former require carefully tuned reward functions and may still often produce unsuitable data, and while the latter more readily produces high-quality data, can be time-intensive to collect. 
The former require carefully tuned reward functions, while the latter can be laborious to collect. Virtual reality~\cite{erickson2020assistive, zhang2018deep} can help, but the human effort remains considerable for complex tasks, and when many demonstrations are required. %for training a deep model to generalize well. 
%By bridging these modes, we may apply past insights from human-in-the-loop shared autonomy~\cite{havoutis2019learning, gopinath2016human, reddy2018shared, jeon2020shared}, to the problem of large-scale data collection for robotic tasks. 
Shared autonomy~\cite{havoutis2019learning, gopinath2016human, reddy2018shared, jeon2020shared} offers a better solution to collecting large-scale data. These works blend robot and user intent using optimization~\cite{gopinath2016human}, reinforcement learning~\cite{reddy2018shared}, and learned coarse-to-fine user precision~\cite{jeon2020shared}, while ours lets the user look far into the future to understand the autonomous prediction. A similar forecasting method was proposed by Liu \textit{et al.}~\cite{liu2020understanding}, but it is used in a behavior cloning loss function, \textcolor{black}{rather than for communicating intent to the user.} 
%Gopinath \textit{et al.}~\cite{gopinath2016human} use a control blending method to share autonomy for assistive reaching tasks using a joystick controller. Reddy \textit{et al.}~\cite{reddy2018shared} blend control using Q-learning to select the a high value action close to the user's input. Jeon \textit{et al.}~\cite{jeon2020shared} introduce method allowing the user to switch to a higher precision shared control based on the robot's confidence of the goal. In contrast to these works, the use of transformers lets the user look far into the future to better understand the autonomous prediction at little computational cost. A similar multi-step look ahead method was proposed by Liu \textit{et al.}~\cite{liu2020understanding}, but it is used in a behavior cloning loss function, rather than for shared autonomy.
We take some insight from P{\'e}rez-D’Arpino and Shah~\cite{perez2017c}, who overlay a series of robot configuration renderings through time to show planned motions. Later, they used this feature for human robot teaming to allow an operator to either accept a suggested motion plan or momentarily intervene~\cite{perez2020experimental}. %Our interface operates similarly to this.

%\subsection{Sequence Modelling for Robotic Manipulation}
Transformers have only recently gained traction in robotics. Janner~\textit{et al.}~\cite{janner2021reinforcement} reframed RL as sequence modeling, and used transformers to control humanoid walking. Chen~\textit{et al.}~\cite{chen2021decision} concurrently explored this in the context of a game environment. %Ours takes influence from these to formulate robot states and actions into the transformer. Using Riemannian motion policies~\cite{ratliff2018riemannian}, we are able to control the robot entirely in end effector space with the transformer, which is more explainable to the human operator and easier to control.
Common transformer implementations have used sinusoidal positional embeddings to better model the order of words~\cite{vaswani2017attention}. However, Chen~\textit{et al.}~\cite{chen2021decision} used an episodic time-step positional embedding and Press~\textit{et al.}~\cite{press2021train} added a linear bias to each attention score. We take inspiration from these and use a cumulative distance embedding to provide information for how far the end effector has traversed. % in Euclidean space. 

\begin{figure*}[t!]
\centering
\includegraphics[width=14cm]{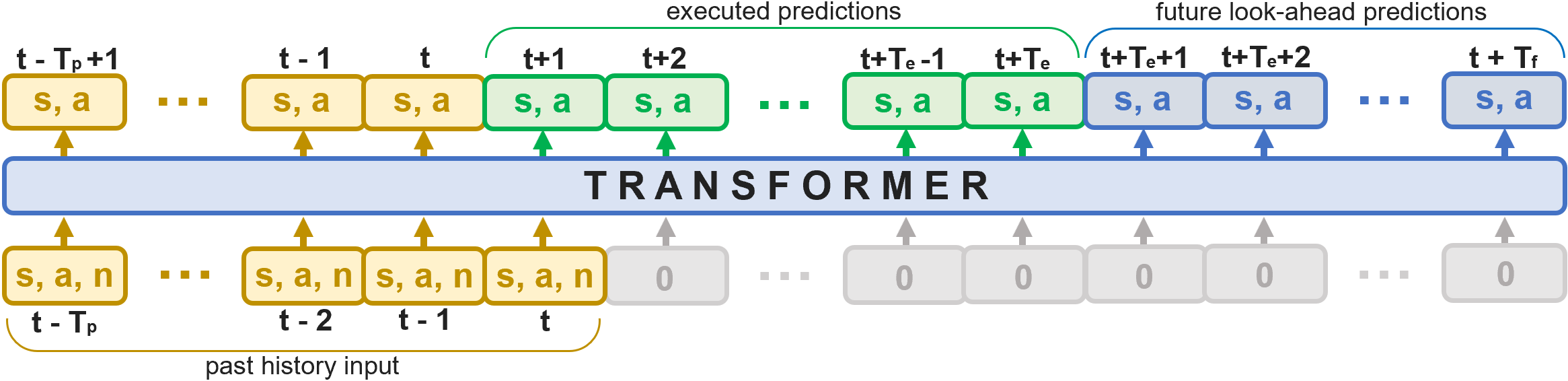}
\vspace{-3mm}
\caption{\footnotesize{The transformer takes as input a past history of $T_p$ states, actions, and positional embeddings, and outputs predicted states and actions. The user can observe predictions far into the future (e.g. $T_f=300$ time-steps) and may choose to execute $T_e <= T_f$ of those predictions. The input for future states is padded with zero, forcing the model to learn future predictions only from past inputs. The total number of time-steps in the sequence is denoted by $T_s$.}}
\label{fig:transformer_arch}
\end{figure*}

\section{Methods}

The transformer takes as input a trajectory of states and actions $\tau_x = \{\mathsf{s}_{\mathsf{x}}, \mathsf{a}_{\mathsf{x}}\}$ and positional encoding vector $\bm{n}$, and outputs a predicted trajectory $\widehat{\tau}_y = \{\mathsf{\widehat{s}}_{\mathsf{y}}, \mathsf{\widehat{a}}_{\mathsf{y}}\}$. As shown in Fig.~\ref{fig:transformer_arch}, the input consists of $T_p$ past time-steps of information, while the output contains  additional information on future predictions up to time-step $T_f$. For each model prediction, the user can choose to execute some number of open-loop actions $T_e$ while observing actions far into the future. 

\subsection{Formulation in Robotics} 
The state at each time-step $\mathsf{s}_\mathsf{t}$ is a vector consisting of a global robot end effector pose $\mathsf{s}_{\mathsf{r},\mathsf{t}} \in \mathbb{R}^7$, continuous gripper state $\mathsf{s}_{\mathsf{g},\mathsf{t}} \in \mathbb{R}^1$, and the local pose of $J$ objects in the environment $\{\mathsf{s}_{\mathsf{o1},\mathsf{t}} \ldots \mathsf{s}_{\mathsf{oJ},\mathsf{t}} \} \in \mathbb{R}^{7 \times J}$ relative to the robot end effector. %as shown in Fig.~\ref{fig:transformer_state}. 
Each 7D pose contains a position and quaternion. At the input, this is fed into a state embedding function $\mathcal{F}_\mathsf{s}$ which we represent with a single fully connected network layer: %$\mathsf{s}_{\textsf{emb}, \mathsf{t}} = \mathcal{F}_\mathsf{s}(\mathsf{s}_\mathsf{t})$, 
\begin{equation}
{\mathsf{s}_{\textsf{emb}, \mathsf{t}}} = \mathcal{F}_\mathsf{s}(\mathsf{s}_\mathsf{t})\label{eq:stateemb}
\end{equation}
where $\mathsf{s}_{\textsf{emb}, \mathsf{t}} \in \mathbb{R}^{128}$. The action at each time-step $\mathsf{a}_\mathsf{t}$ is a vector consisting of the local target position of the robot end effector $\mathsf{a}_{\mathsf{r},\mathsf{t}} \in \mathbb{R}^7$ and the binary gripper command $\mathsf{a}_{\mathsf{g},\mathsf{t}} \in \mathbb{R}^1$. At the input, this is fed into an action embedding function $\mathcal{F}_\mathsf{a}$, which we represent with a single fully connected network layer: %$\mathsf{a}_{\textsf{emb}, \mathsf{t}} = \mathcal{F}_\mathsf{a}(\mathsf{a}_\mathsf{t})$,
\begin{equation}
{\mathsf{a}_{\textsf{emb}, \mathsf{t}}} = \mathcal{F}_\mathsf{a}(\mathsf{a}_\mathsf{t})\label{eq:actionemb}
\end{equation}
where $\mathsf{a}_{\textsf{emb}, \mathsf{t}} \in \mathbb{R}^{128}$. Additionally, the network contains a positional embedding to measure the distance and rotation the end effector has traversed at each time-step along the trajectory from time-step $1 \ldots t$. The positional embedding $\mathsf{n}_\mathsf{t} \in \mathbb{R}^1$ is an integer token computed at each time-step as: %$\mathsf{n}_\mathsf{t} = \sum_{t^{\prime}=2}^t \sum_{j=1}^8 ||\mathsf{c}_{\mathsf{r},\mathsf{j},\mathsf{t}^{\prime}} - \mathsf{c}_{\mathsf{r},\mathsf{j},\mathsf{t}^{\prime}-1}||_2$
%\begin{equation}
%n_t = \sum_{i=2}^t ||p_{R,i} - p_{R,i-1}||_2 + \lambda \sum_{i=2}^t ||q_{R,i} - q_{R,i-1}||_2
%\end{equation}
% \vspace{-5mm}
 \begin{equation}
 \mathsf{n}_\mathsf{t} = \sum_{t^{\prime}=2}^t \sum_{j=1}^8 ||\mathsf{c}_{\mathsf{r},\mathsf{j},\mathsf{t}^{\prime}} - \mathsf{c}_{\mathsf{r},\mathsf{j},\mathsf{t}^{\prime}-1}||_2
 \end{equation}\label{eq:pos_enc}
% \vspace{-5mm}
where each $\mathsf{c}_{\mathsf{r}}$ is coordinate of a corner of a 3D bounding box around the end effector at a given time-step, and is a function of end effector position $\mathsf{p}_\mathsf{r}$ quaternion $\mathsf{q}_\mathsf{r}$ in global frame. This pose representation is the same as that in the loss function by Allshire \textit{et al.}~\cite{allshire2021transferring}. We chose this pose representation because it casts rotation in position space, which mitigates the problem of combining heterogenous terms in the same function. The token $\mathsf{n}_\mathsf{t}$ is fed into a learned positional embedding layer, $\mathcal{F}_\mathsf{n}$: %$\mathsf{n}_{\textsf{emb},\mathsf{t}} = \mathcal{F}_\mathsf{n}(\mathsf{n}_\mathsf{t})$, 
\begin{equation}
    \mathsf{n}_{\textsf{emb},\mathsf{t}} = \mathcal{F}_\mathsf{n}(\mathsf{n}_\mathsf{t})
\end{equation}

where $\mathsf{n}_{\textsf{emb},\mathsf{t}} \in \mathbb{R}^{128}$. This is added to each state and action embedding. At each time-step the transformer receives the input vector computed as %$\mathsf{x}_\mathsf{t} = \textsf{LN}\big((\mathsf{s}_{\textsf{emb},\mathsf{t}} + \mathsf{n}_{\textsf{emb},\mathsf{t}}) \oplus (\mathsf{a}_{\textsf{emb},\mathsf{t}} + \mathsf{n}_{\textsf{emb},\mathsf{t}})\big)$
 \begin{equation}
 \mathsf{x}_\mathsf{t} = \textsf{LN}\big((\mathsf{s}_{\textsf{emb},\mathsf{t}} + \mathsf{n}_{\textsf{emb},\mathsf{t}}) \oplus (\mathsf{a}_{\textsf{emb},\mathsf{t}} + \mathsf{n}_{\textsf{emb},\mathsf{t}})\big)
 \end{equation}
 
where $\mathsf{x}_\mathsf{t} \in \mathbb{R}^{256}$ and $\textsf{LN}$ represents layer normalization. The transformer outputs predicted vector $\widehat{\mathsf{y}}_\mathsf{t} \in \mathbb{R}^{256}$, which is then decoded with linear layers on the output mirroring those on the input, represented by $\mathsf{g}_\mathsf{s}$ and $\mathsf{g}_\mathsf{a}$ for the states and actions. The output of these decoding layers contain predicted states and actions, the sequence of which forms trajectory $\widehat{\tau}_y$. The robot is controlled with the predicted end-effector pose actions by using Riemannian motion policies~\cite{ratliff2018riemannian}. 

\subsection{Network Training} During training, a sub-sequence of length $T_s = 400$ from is sampled from a task demonstration of trajectory length $T_d$, where $T_d > T_s$ (recall Fig. 2). The network takes input of sequence length $T_s$ of state and action pairs, $\tau_x$, associated with each time-step. Inspired by BERT~\cite{devlin2018bert} masking, we only keep the inputs from the first $T_p$ time-steps and mask the remaining inputs from time-steps $T_s - T_p$ with zeros. The network is trained to predict the corresponding state and action pairs, $\tau_y$, at masked out time-steps.
%Data from the first $k$ time-steps is allocated to input $\tau_x$, where $T_p$ is the length of the input time series and $T_p < T_s$. Data of length $T_s$ from the entire sequence are allocated to ground truth label $\tau_y$. To fit $\tau_x$ of length $T_p$ into a transformer of length $T_s$, the last $T_s-T_p$ time-steps are masked with zero. 
The input length $T_p$ is chosen at random during training to force the transformer to make future predictions of an arbitrary horizon length. It is sampled from a uniform distribution $T_p \sim \mathcal{U}(1, 350)$, such that the future prediction length is at least 50 time-steps. 

% We define the each trajectory state $\bm{s}$
% Loss function for decision transformers 

We denote $\mathsf{s}_{\mathsf{r}, \mathsf{t}} = [\mathsf{p}_{\mathsf{r},\mathsf{t}}, \mathsf{q}_{\mathsf{r},\mathsf{t}}]$ to be the state of the robot end-effector composed of the position and orientation at time $\mathsf{t}$. We denote $\mathsf{a}_\mathsf{t}^{\mathsf{r}} = [\mathsf{p}_{\mathsf{r_{target}}, \mathsf{t}}, \mathsf{q}_{\mathsf{r_{target}}, \mathsf{t}}]$ to be the robot action composed of the target end-effector position and orientation. Similarly, we define the current gripper state $\mathsf{s}_{\mathsf{g},\mathsf{t}} \in [0, 1]$ and the target gripper state $\mathsf{a}_{\mathsf{g},\mathsf{t}} \in [0, 1]$. Object $\mathsf{i}$ in the scene is also represented by its state vector $\mathsf{s}_{\mathsf{oi},\mathsf{t}} = [\mathsf{p}_{\mathsf{oi},\mathsf{t}}, \mathsf{q}_{\mathsf{oi},\mathsf{t}}]$, which is composed of its position and orientation. The network takes an input sequence of states $\mathsf{s}_{\mathsf{x}}$=$\{\mathsf{s}_{\mathsf{r},\mathsf{t}}, \mathsf{s}_{\mathsf{o1},\mathsf{t}} \cdots \mathsf{s}_{\mathsf{oJ},\mathsf{t}}, \mathsf{s}_{\mathsf{g},\mathsf{t}} \}_{\mathsf{t}=0}^{\mathsf{k}}$ composed of the robot end-effector state, gripper state and states of $\mathsf{J}$ objects in the scene, as well as the actions $\mathsf{a}_{\mathsf{x}}$=$\{\mathsf{a}_{\mathsf{r},\mathsf{t}}, \mathsf{a}_{\mathsf{g},\mathsf{t}} \}_{\mathsf{t}=0}^{\mathsf{k}}$ associated with the robot end-effector. It then predicts the future sequence of states $\mathsf{s}_{\mathsf{y}}$=$\{\mathsf{s}_{\mathsf{r},\mathsf{t}}, \mathsf{s}_{\mathsf{o1},\mathsf{t}} \cdots \mathsf{s}_{\mathsf{oJ},\mathsf{t}}, \mathsf{s}_{\mathsf{g},\mathsf{t}} \}_{\mathsf{t=k+1}}^{\mathsf{T}}$ and corresponding actions $\mathsf{a}_{\mathsf{y}}$=$ \{\mathsf{a}_{\mathsf{r},\mathsf{t}}, \mathsf{a}_{\mathsf{g}, \mathsf{t}} \}_{\mathsf{t=k+1}}^{\mathsf{T}}$. We predict the future states and actions given the current states and actions using a GPT-style transformer: $\mathsf{\widehat{s}}_{\mathsf{y}}, \mathsf{\widehat{a}}_{\mathsf{y}} = \mathsf{GPT}(\mathsf{s}_{\mathsf{x}}, \mathsf{a}_{\mathsf{x}})$.

\textbf{Loss function.} The loss function used to train the transformer model has a number of components that enable it to learn states and actions. We compute losses related to the end-effector state and action, the gripper state and action, and to the state of each object in the scene. 

The state and action space is composed of positions and orientations, which are heterogeneous terms that would require a weighting factor if directly combined in the loss function. Instead, we compute the loss in euclidean space in a similar way to Eq. 3, because it casts position and orientation into 3D locations of 8 bounding box corners around the end effector or object. We compute the end-effector state loss by mapping its position and orientation to the box corners for both the predictions and ground truth: 

\begin{eqnarray}
\widehat{\mathsf{c}}_{\mathsf{s}_\mathsf{r}} = \textsf{CORNERS}(\widehat{\mathsf{p}}_{\mathsf{s}_\mathsf{r}}, \widehat{\mathsf{q}}_{\mathsf{s}_\mathsf{r}}) \\ 
\mathsf{c}_{\mathsf{s}_\mathsf{r}} = \textsf{CORNERS}(\mathsf{p}_{\mathsf{s}_\mathsf{r}}, \mathsf{q}_{\mathsf{s}_\mathsf{r}})
\end{eqnarray}

where $\mathsf{c}_{\mathsf{s}_\mathsf{r}} \in \mathbb{R}^{8 \times 3}$ are the 3D positions of the 8 corners of the bounding box extents of the end effector. The loss is computed as the $L_2$ distance of the corresponding 8 corners of the predictions and ground truth:

\begin{equation}
    \mathcal{L}_{\mathsf{s}_\mathsf{r}} = \sum_{t=T_p}^{T_s} ||\widehat{\mathsf{c}}_{\mathsf{s}_\mathsf{r}, \mathsf{t}} - \mathsf{c}_{\mathsf{s}_\mathsf{r},\mathsf{t}}||_2
\end{equation}

Similarly, we can define a loss on the action space where the predictions are end-effector target positions and orientations, as well as a loss on the state predictions of the objects.

To compute loss on the gripper state and action we use binary cross entropy:
\begin{eqnarray}
    \mathcal{L}_{\mathsf{s}_\mathsf{g}} = \sum_{t=T_p}^{T_s} \textsf{BCE}(\widehat{\mathsf{s}}_{\mathsf{g}, \mathsf{t}}, \mathsf{s}_{\mathsf{g}, \mathsf{t}}) \\ 
    \mathcal{L}_{\mathsf{a}_\mathsf{g}} = \sum_{t=T_p}^{T_s} \textsf{BCE}(\widehat{\mathsf{a}}_{\mathsf{g}, \mathsf{t}}, \mathsf{a}_{\mathsf{g}, \mathsf{t}})
\end{eqnarray}
The total loss is sum of these components:
\begin{equation}
    \mathcal{L}_{\textsf{TOTAL}} = \mathcal{L}_{\mathsf{s}_\mathsf{r}} + \mathcal{L}_{\mathsf{a}_\mathsf{r}} + \sum_{i=1}^{J}\mathcal{L}_{\mathsf{s}_\mathsf{oi}} +  \lambda(\mathcal{L}_{\mathsf{s}_\mathsf{g}} + \mathcal{L}_{\mathsf{a}_\mathsf{g}})
\end{equation}

%\vspace{-3mm}
\section{Evaluation}
%\vspace{-3mm}

We evaluated the transformer and Assistive Tele-op system across a variety of tasks representing industrial~\cite{mandlekar2018roboturk, zeng2020transporter}, household~\cite{chebotar2019closing, garrett2020online}, and caregiving~\cite{grice2019home, erickson2020assistive} task scenarios. %For this, we used both existing data from Roboturk~\cite{mandlekar2018roboturk} and new data that we collected with VR. 
For this, we used both existing data from the Roboturk~\cite{mandlekar2018roboturk} simulation dataset covering pick-and-place and nut assembly tasks, as well as new data for other tasks that we collected in VR.

\subsection{Data collection}
\textbf{Existing data - Roboturk.} The Roboturk simulation dataset was collected in the Mujoco~\cite{todorov2012mujoco} simulator with a Baxter robot, but we found that many demonstrations played back successfully in the NVIDIA Omniverse simulator with a Franka robot when controlling the robot in end effector space using RMPs~\cite{ratliff2018riemannian}. This dataset consists of over 6000 crowd-sourced human demonstrations for pick-and-place tasks with 4 objects (cereal box, milk jug, bread, and coke can) and nut assembly tasks. Of these, we selected 533 pick-and-place demonstrations for pretraining the transformer. We chose the first 533 demonstrations that had an overall time of less than 900 time-steps, because we observed that shorter demonstrations had better quality. We hand-selected demonstrations for the round nut assembly (Task B) by choosing the first 50 that played back smoothly in our recreation of the scene in Omniverse with the Franka robot.

\textbf{New data.} An HTC Vive virtual reality headset was used by a researcher to collect data in Omniverse with Franka. To create variation in each scene, we sampled initial scene object poses from the following uniform noise distributions for both training and testing scenes: (A) block stacking - blue picked block from $\pm 5$cm planar translation and $\pm 45^{\circ}$ rotation, orange stacked block from $\pm 5$cm planar translation. (C) assembly kit - pink hexagon from $\pm 7.5$cm planar translation and $\pm 180^{\circ}$ rotation. (D) cabinet, $\pm 45^{\circ}$ rotation. (E) put bowl in cabinet, cabinet from  $\pm 45^{\circ}$ rotation, green bowl from $\pm 10$cm uni-directional translation relative to the cabinet. (F) itch scratching, scratch tool from $\pm 5$ cm unidirectional translation, humanoid root pose from $\pm 5$ cm planar translation, and 19 unique itch scratching locations on the humanoid. 3 manual demonstrations are collected for each location (57 total). (G) humanoid drinking, mug w/straw from $\pm 5$cm unidirectional translation and $\pm 30^{\circ}$ rotation over the table, and humanoid/wheelchair root from $\pm 7.5$cm planar translation.

\subsection{Transformer evaluation}

The input sequence length was set to $T_p = 250$ and the future prediction length to $T_f = 150$. Each time a forward pass runs on the transformer, the simulator executes $T_e = 10$ actions, which are fed back into the transformer. Predicted actions are fed directly back to the transformer, while real simulator states resulting from the actions are fed into the model input. The transformer contains 6 layers, 8 heads, a hidden layer size of 256, and it is trained with a batch size of 128. During pre-training, we used a learning rate of 1e-4 that linearly decreased to 5e-5, and a learning rate of 5e-5 for training on other task data.

%\subsection{Automatic model prediction}\label{ssec:auto_pred}
%\vspace{-1mm}
\textbf{Automatic model prediction.} First, we evaluated the pretrained model. We trained it for $> 2$ days on \textcolor{black}{raw RoboTurk data with the Baxter robot in MuJoCo, and evaluated it in a reconstructed environment with the Franka robot in NVIDIA Omniverse. We evaluated the pretrained model on both 50 scenes from the training data and on 50 test scenes. The training data evaluation provides a measure of the sim2sim transfer between simulators. The test data evaluation shows generalization to previously unseen initial item locations. We conducted more tests in tasks A-G in Omniverse (see Table~\ref{tbl:success}).} For each, we fine-tuned a transformer from the pretrained model and trained a transformer from scratch (no pretraining). We used a fixed-time budget for this comparison. %The number of training demonstrations for each task are stated in Table~\ref{tbl:success}. 
Each model is tested on 50 new object configurations, except itch scratching, which is tested on 57. %Evaluation examples are shown in Fig.~\ref{fig:results_success}.

\begin{table*}[t!]
\begin{center}
\footnotesize
\caption{\footnotesize{Success rate and demonstration time for models trained from scratch and pretrained models.} %Automatic prediction and assistive tele-op metrics are evaluated using previously unseen task scenarios. 
%During assistive tele-op, a human user momentarily intervenes when the model prediction forecast veers off course.
}\label{tbl:success}
\renewcommand{\arraystretch}{1.1}
\vspace{1mm}
\scalebox{0.83}{
\begin{tabular} {l|c||c|c|c|c||c|c|c}
 & No. &  \multicolumn{4}{c||}{Success rate} & \multicolumn{3}{c}{Manual demonstration time (s)} \\
\cline{3-9}
& training & Manual & Auto & Auto & Assistive & Manual & Auto & Assistive \\
Task & demos & tele-op & no pretr. & w/pretr. & Tele-op & tele-op &  & Tele-op \\
\hline
\cline{1-9}
RoboTurk pick/place~\cite{mandlekar2018roboturk} &  533 & - & \hspace{-1mm}0.84 / 0.66$^\dagger$\hspace{-1mm} & - & - & - & - & -\\
\hline
\hline
A. Block stacking & 35 & 1.00 & 0.60 & 0.74 & 1.00 & 14.5 & 0.0 & 3.5\\
\hline
B. Round nut assembly~\cite{mandlekar2018roboturk} & 50 & 1.00 & 0.38 & 0.70 & 0.94 & 72.3*\hspace{-1.5mm} & 0.0 & 7.9 \\
\hline
C. Assembly kit - hexagon~\cite{zeng2020transporter} & 35 & 1.00 & 0.00 & 0.10 &  0.92 & 38.5 & 0.0 & 5.6\\
\hline
D. Cabinet drawer opening & 35 & 1.00 & 0.98 & 1.00 & 1.00 & 14.0 & 0.0 & N/A \\
\hline
E. Put bowl in drawer & 50 & 1.00 & 0.52 & 0.64 & 1.00 & 28.3 & 0.0 & 4.5\\
\hline
% EE. Put bowl in drawer, +/- 5cm & 50 &  & & & & & \\
% \hline
F. Humanoid itch scratching & 57 & 1.00 & 0.46 & 0.70 & 0.93 & 23.0 & 0.0 & 4.2\\
\hline
G. Humanoid drinking & 35 & 1.00 & 0.68 & 0.82 & 0.94 & 18.8 & 0.0 & 7.7\\
\hline
Overall (A-G average) & - & \textbf{1.000} & \textbf{0.517} & \textbf{0.671} & \textbf{0.961}  & \textbf{29.9} & \textbf{0.0} & \textbf{5.5} \\
\hline
\cline{1-9}
\multicolumn{9}{l}{\scriptsize *\textit{Based on approximate RoboTurk data frequency of 15 Hz.} $^{\dagger}$\textit{Results on training data / results on test data. All results collected in NVIDIA Omniverse with Franka.}}
\end{tabular}}
\end{center}
\vspace{-3mm}
\end{table*}

%\vspace{-3mm}
%\subsection{Assistive teleoperation}
%\vspace{-1mm}
\textbf{Assistive Tele-op:} We evaluated the human-in-the-loop Assistive Tele-op system using both task success rate and manual demonstration time elapsed. A researcher used the HTC Vive VR system to communicate with the transformer prediction when controlling the robot, as shown in Fig.~\ref{fig:interventions}. For each task, the model began in automatic mode, and the user clicked a button on the interface to stage an intervention when the robot moved in an inappropriate direction (e.g. away from the bowl rather than toward it when picking it up). We score Assistive Tele-op success as the ability to complete a demonstration on a new task scenario -- either with fully automatic prediction, or with intervention assistance. Assistive Tele-op can only improve the demonstration success rate. For all Assistive Tele-op scenarios, we used the transformer with pretraining. Time taken for manual demonstration is compared among manual, automatic, and assistive modes. In manual mode, this is the average time to collect each full demonstration. In automatic, it is 0, because no human effort is required. In Assistive Tele-op mode, it is the average intervention time for all demonstrations per task.

%\vspace{-3mm}
\section{Results and Discussion}
%\vspace{-3mm}

% \textbf{The pre-trained model performs well.} We tested the pretrained model performance on unseen examples from the test set, and it achieves good performance of 66\%.

\begin{figure*}[t!]
\centering
\includegraphics[width=14cm]{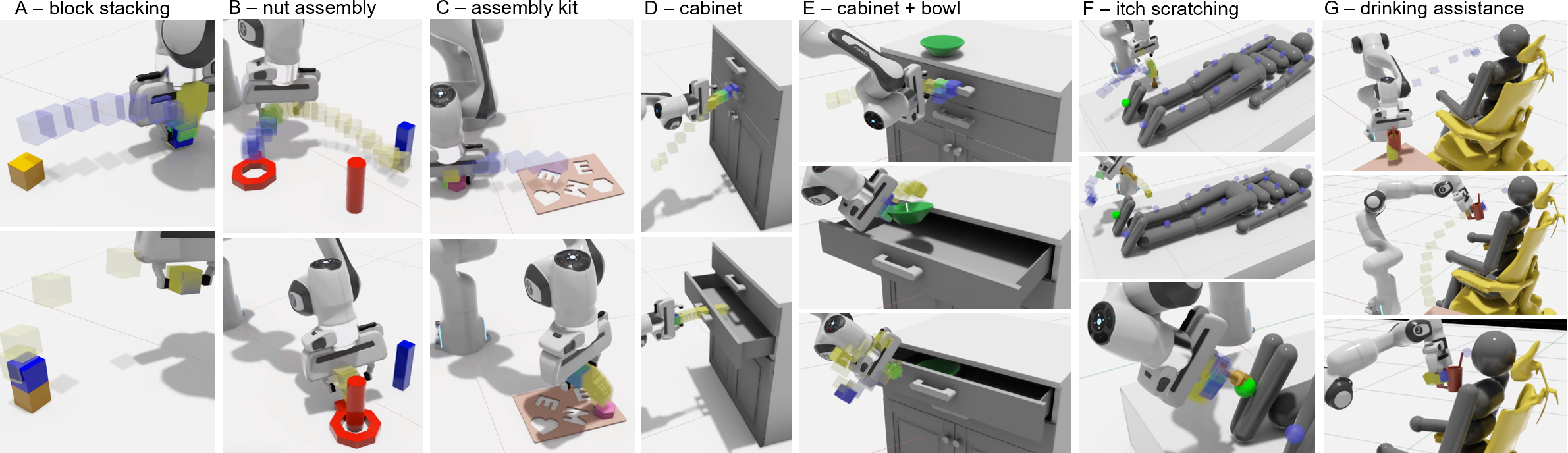}
\vspace{-6mm}
\caption{\footnotesize{Fully automatic prediction task examples that the transformer completes successfully. (A) the blue block is stacked upon the orange block (B) the round nut is placed on the round peg. (C) the pink hexagon is fit into the assembly kit board. (D) the bottom cabinet drawer is opened. (E) the green bowl is put in the top cabinet drawer. (F) an itch scratching tool is picked up and used to scratch an itch on the bottom of the left foot. (G) a mug with a straw is picked and brought to a person in a wheelchair.}}
\vspace{-3mm}
\label{fig:results_success}
\end{figure*}

\textbf{Pre-trained transformers can be used for few shot generalization to new tasks.} For each task, we compared models trained from scratch to those fine-tuned starting with a pre-trained model. The pre-trained and fine-tuned models performed better across all tasks. Success examples with fully automatic model prediction are shown in Fig.~\ref{fig:results_success}. Testing scenarios were successful in most cases, except Task C (see section~\ref{ssec:limitations}.)

%\vspace{-1mm}
\textbf{Models trained with our method can transfer between different simulators and robots.} The Roboturk dataset was collected in Mujoco using a Baxter robot. However, we tested the transformer model using a Franka robot in the Omniverse~\cite{nvidia2021} %\footnote{\url{https://developer.nvidia.com/nvidia-omniverse-platform}} 
simulator. These environments have a different robot configuration, control method, data collection rate, and simulation method, but the model performs well, showing good sim2sim transfer. Task B, also from Roboturk, provides further evidence for this. Formulating the model in end effector space is key to this transfer, by obviating the configuration space representation that is different between Baxter and Franka.

\begin{figure}[t!]
\centering
\vspace{-3mm}
\includegraphics[width=14cm]{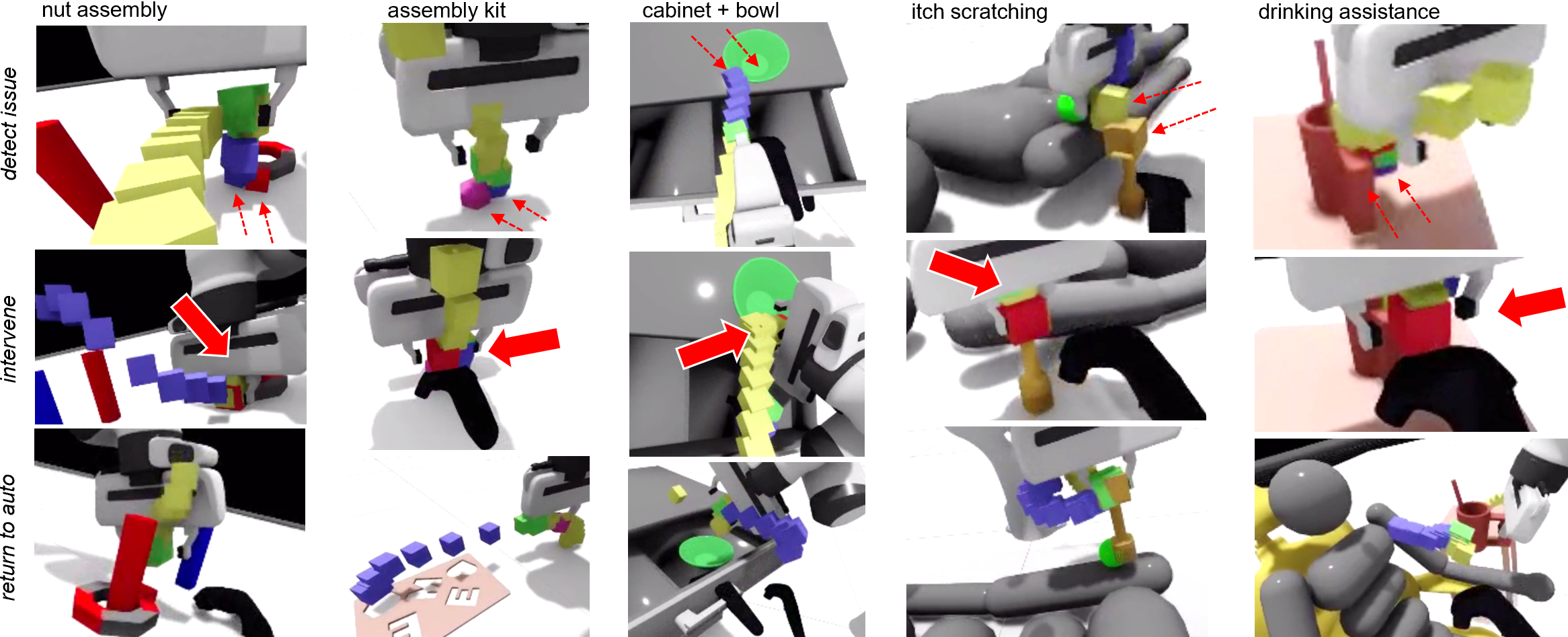}
\vspace{-5mm}
\caption{\footnotesize{Assistive Tele-op in VR. When the user detects future actions that are inappropriate, they click a button to take over control. After a momentary nudge, they return control to the transformer.}}
\vspace{-4mm}
\label{fig:interventions}
\end{figure}

%\vspace{-1mm}
\textbf{Human interventions can get the model back on track.} When a human intervenes in event of failure to nudge the robot back on track, success increases from 67.1\% to 96.1\%. See Fig.~\ref{fig:interventions}.

%\vspace{-1mm}
\textbf{Collecting Assistive Tele-op demonstrations with model prediction is easier.} For purely manual teleoperation, the average demonstration time is 29.9 seconds. For Assistive Tele-op, the average human demonstration time to get the robot back on track is 5.5 seconds across task scenarios that otherwise cannot be completed with fully automatic model prediction.

%\vspace{-1mm}
\textbf{Auxiliary object pose loss boosts performance.} We ablated the auxiliary loss on the objects in the scene, and found that for the round nut assembly (Task B) success for the pretrained model, performance drops from 70\% to 58\%.

\subsection{Limitations}\label{ssec:limitations} The transformer model has poor generalization performance for precision tasks with few (<= 50) examples. The Assembly Kit from TransporterNets consists of five precisely fitting shapes into a board. The performance is low for a single precisely fitting shape (the Hexagon, at 10\%), as shown in Table~1. However, if the goal criteria is set more loosely (i.e. the shape is next to the goal but not quite in the slot), then performance is 46\%. It also has some difficulty picking the hexagon, because the hexagon is almost as wide as the open gripper max width.

Most failures happen due to imprecise grasping when robot is unable to recover after failing at the first attempt. In some cases, the robot grasps the object but stops midway and never reaches the goal or stays frozen after grasping. This may be to due to out of distribution errors as a result of limited demonstrations.

\small

\bibliographystyle{plain}
\bibliography{main}

\end{document}